\documentclass{article}
\usepackage{spconf,amsmath,graphicx}
\usepackage{subcaption}
\usepackage{amssymb}
\usepackage{caption}
\usepackage{multirow}
\usepackage{color}
\usepackage[colorlinks,
            linkcolor=red,
            anchorcolor=blue,
            citecolor=green
            ]{hyperref}

\title{SEMANTIC SEGMENTATION IN LEARNED COMPRESSED DOMAIN}
\name{Jinming Liu$^*$ \qquad Heming Sun$^1$$^2$ \qquad Jiro Katto$^*$$^1$}
\address{$^*$Department of Computer Science and Communication Engineering, Waseda University, Tokyo, Japan\\
$^1$Waseda Research Institute for Science and Engineering, Waseda University, Tokyo, Japan\\
$^2$JST PRESTO, 4-1-8 Honcho, Kawaguchi, Saitama, Japan}

\begin{document}
%
\maketitle
\begin{abstract}
Most machine vision tasks (e.g., semantic segmentation) are based on images encoded and decoded by image compression algorithms (e.g., JPEG). However, these decoded images in the pixel domain introduce distortion, and they are optimized for human perception, making the performance of machine vision tasks suboptimal. In this paper, we propose a method based on the compressed domain to improve segmentation
tasks. i) A dynamic and a static channel selection method are proposed to reduce the redundancy of compressed representations that are obtained by encoding. ii) Two different transform modules are explored and analyzed to help the compressed representation be transformed as the features in the segmentation network. The experimental results show that we can save up to 15.8\% bitrates compared with a state-of-the-art compressed domain-based work while saving up to about 83.6\% bitrates and 44.8\% inference time compared with the pixel domain-based method.
\end{abstract}
\section{Introduction}
\label{sec:intro}
With the development of deep learning, machine vision tasks have been widely implemented in the past few years and have played important roles in our lives. 
For example, semantic segmentation is applied to medical image analysis, autonomous vehicles, video surveillance, augmented reality, and many other areas \cite{minaee2021image}. However, many machine vision tasks are usually based on decoded images that need image compression.The distortion of decoded images deteriorate
 the performance of machine vision tasks.
\par

Classical image compression algorithms, such as, JPEG \cite{wallace1992jpeg}, and HEVC/H.265-intra \cite{sullivan2012overview} have achieved good Rate-distortion performance in human perception while some learned end-to-end image compression methods \cite{balle2018variational, cheng2020learned} also achieved comparable performance. However, all of them are designed to reduce the distortion which is perceived by human and has a gap with machine perception. Therefore, current algorithms usually lead to suboptimal semantic segmentation. 

\par
\begin{figure}[ht]
	\centering
	\includegraphics[scale=0.7]{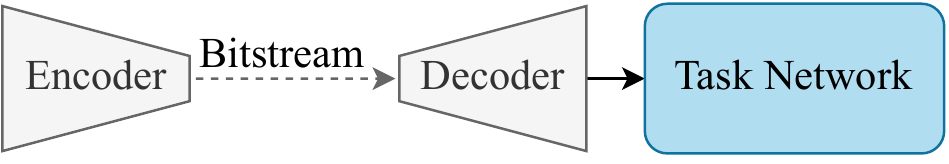}
	\caption{The pixel domain-based method.
	}
	\label{fig:20}
\end{figure}

Some methods based on the compressed domain are proposed to solve these problems. Pixel domain-based methods mean we need to input the images encoded and decoded by an image compression algorithm to the semantic segmentation network, as Fig. \ref{fig:20} shows. Different from that, compressed domain-based methods mean we directly input the representations in the compressed domain into the semantic segmentation network without the computation cost for the decoded image. A related standard named JPEG AI \cite{JPEGAI} had been proposed. Torfason $et\ al.$ \cite{torfason2018towards} had proved that the compressed domain-based method could achieve a better performance than the pixel domain-based method, and made some experiments on image classification and object detection. Choi $et\ al.$ \cite{choi2021scalable} further verified this by using an information theory method and proposed a compression domain-based approach which is verified on object detection and semantic segmentation. Bai $et\ al.$ \cite{Bai2022AAAI} utilized a transformer to improve image classification based on compressed domain. 
Wang $et\ al.$ \cite{wang2022learning} proposed a channel selection method and an entropy estimation method to improve the results in \cite{torfason2018towards}. However, the channels selection method can only delete the channels with the lowest entropy since the selection method is based on the variance of the compressed representations.
\par
In this paper, we propose a method based on the compressed domain to improve segmentation tasks.
The main contributions of this paper are as follows:
\begin{itemize}

    \item Based on a learned gate module, a dynamic and a static channel selection method are proposed to select suitable channels and reduce the redundancy of compressed representations.

    \item We explore and analyze two different modules transforming the representation in the compressed domain to the feature that comes from the mid-layer in the segmentation network.

    \item We perform experiments based on Cityscapes, and the results show that, in the same mIoU cases, we can save up to 15.8\% bitrates compared with a state-of-the-art compressed domain-based work while saving up to about 83.6\% bitrates and 44.8\% inference time compared with the pixel domain-based method.
\end{itemize}

\section{Method}
\begin{figure*}[ht]
	\centering
	\includegraphics[scale=0.6]{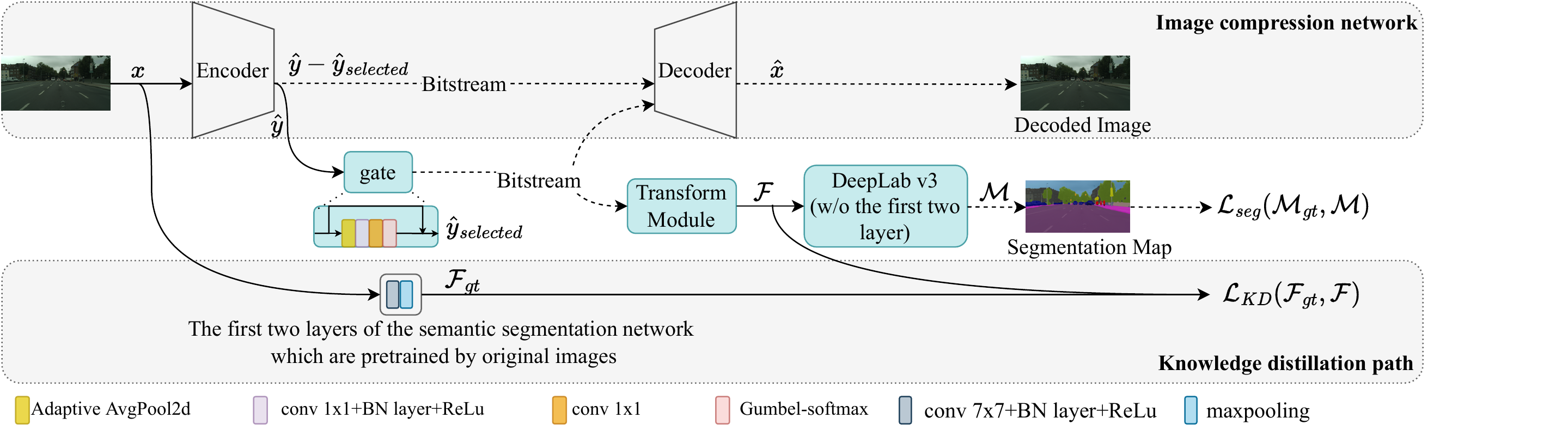}
	\caption{The framework of our method (The knowledge distillation path is firstly trained by original images individually. Both the weights of this path and the pretrained weights of the compression network are frozen during the training.).
	}
	\label{fig:1}
\end{figure*}
\subsection{Framework}

The framework of our method is shown in Fig. \ref{fig:1}. It consists of five modules: an image compression network, a gate module, a transform module, a segmentation network, and a knowledge distillation path. The hyperprior model in \cite{balle2018variational} is set as the image compression network whose the pre-trained weights are provided by CompressAI \cite{begaint2020compressai}. In the following steps, the weights of this compression network are frozen so that our method will not affect the quality (e.g, PSNR) of the decoded images.

For semantic segmentation network, we choose DeepLab v3 \cite{chen2017rethinking} whose first two layers are removed to conform to the size of the compressed representation. And then, to gain a better performance, a knowledge distillation method is utilized. The teacher network of knowledge distillation is shown in the knowledge distillation path that contains the first two layers of DeepLab v3, which are pretrained by raw images, and are frozen after that. The student network is the all networks above the knowledge distillation path.
\par 
We firstly input the raw image $x$ to the encoder of the compression network to get a compressed representation $\hat{y}$. For the pixel domain-based method, we need to transmit this representation $\hat{y}$ to the decoder to get the decoded image $\hat{x}$, and input the $\hat{x}$ to the semantic segmentation network to complete segmentation. But for our compressed domain-based method, to get a machine vision task result, we do not need the decoder to get $\hat{x}$, and just input $\hat{y}$ to the gate module to get a selected representation $\hat{y}_{selected}$.

And then, the transform module is used for transforming $\hat{y}_{selected}$ to the feature $\mathcal{F}$ for the segmentation network. The feature is further input to semantic segmentation network to get the segmentation map $\mathcal{M}$. A cross-entropy loss $\mathcal{L}_{seg}$ of $\mathcal{M}$ and ground truth $\mathcal{M}_{gt}$ is calculated for updating the network. We can express it as:
\begin{equation}
    \mathcal{L}_{seg} = \mathbb{S G}(CrossEntropy(\mathcal{M}, \mathcal{M}_{gt}); \phi)
\end{equation}
where $\mathbb{S G}$ is the stop-gradient operator since the compression network parameters $\phi$ are not updated during the training.
\par
To further improve the performance, another knowledge distillation loss $\mathcal{L}_{KD}$ is calculated by the knowledge distillation path. We input $x$ into the frozen first two layers of the semantic segmentation network to get the $\mathcal{F}_{gt}$. Then we calculate the mean square error of $\mathcal{F}_{gt}$ and $\mathcal{F}$ as $\mathcal{L}_{KD}$, which can be expressed as:
\begin{equation}
    \mathcal{L}_{KD} = \mathbb{S G}(MSE(\mathcal{F}, \mathcal{F}_{gt}); \psi)
\end{equation}
where $\psi$ means the weights of the knowledge distillation path, and these weights are frozen during the training.
\par
At last, the total loss $\mathcal{L}$ can be fomulated as:
\begin{equation}
    \mathcal{L} = \mathcal{L}_{seg} + \lambda_1 \mathcal{L}_{KD}
\end{equation}
where $\lambda_1$ is used for controlling the weights between $\mathcal{L}_{KD}$ and  $\mathcal{L}_{seg}$, here we set it as 1 based on auxiliary experiments.
\par On the other hand, by continuing to transmit the residuals between $\hat{y}$ and $\hat{y}_{selected}$, we can also get a decoded image. 

\subsection{Channel Selection}
Previous works \cite{choi2021scalable, wang2022learning, liu2021} have shown that there are some channels in the compressed representations that are useless for machine vision tasks. Here we propose a gate module to adaptively learn which channel is suitable for the semantic segmentation task. On this basis, dynamic selection and static selection methods are proposed.
\begin{figure}[ht]
	\centering

	\includegraphics[scale=0.6]{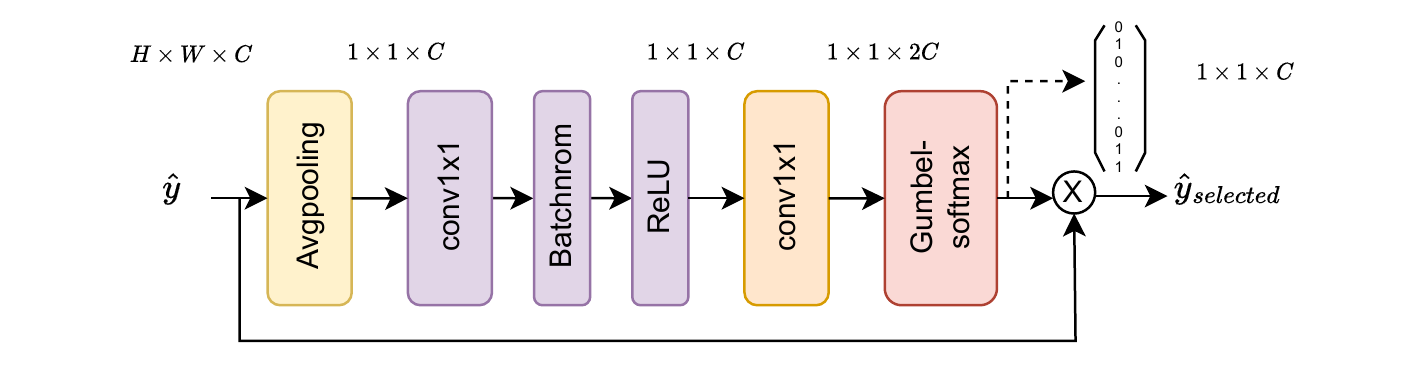}
	\caption{The framework of the gate module.
	}
	\label{fig:10}
\end{figure}
\begin{figure}
\captionsetup[subfigure]{labelformat=empty}
  \centering


 \begin{subfigure}[htb]{.5\linewidth}
    \centering\includegraphics[width=0.9\linewidth]{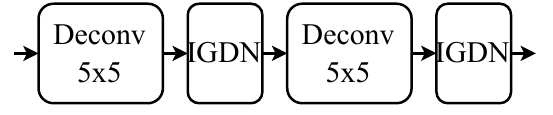}
    \caption{(a) Module-Deconv}
  \end{subfigure}
  
  \begin{subfigure}[htb]{1.\linewidth}
    \centering\includegraphics[width=1.\linewidth]{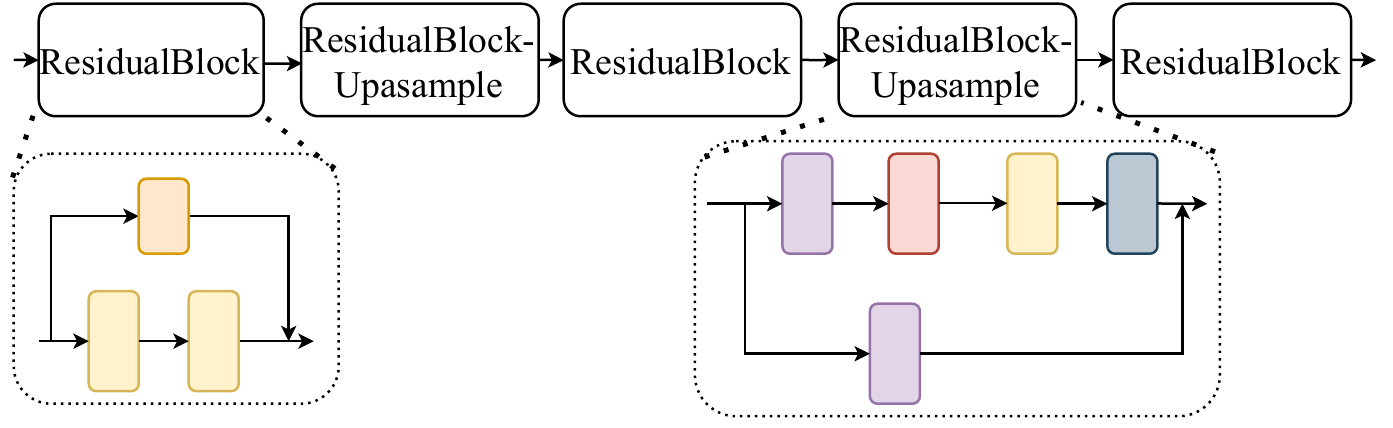}
    \caption{(b) Module-Res}
  \end{subfigure}

  \begin{subfigure}[htb]{1.\linewidth}
    \centering\includegraphics[width=1.\linewidth]{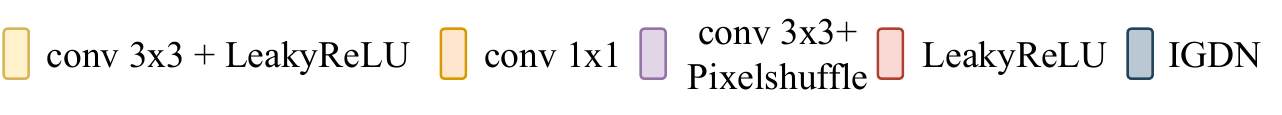}
  \end{subfigure}
  \caption{The two different transform modules.}
  \label{fig:2}

\end{figure}

\begin{figure*}
\captionsetup[subfigure]{labelformat=empty}
  \centering


 \begin{subfigure}[ht]{.24\linewidth}
    \centering\includegraphics[width=1.\linewidth]{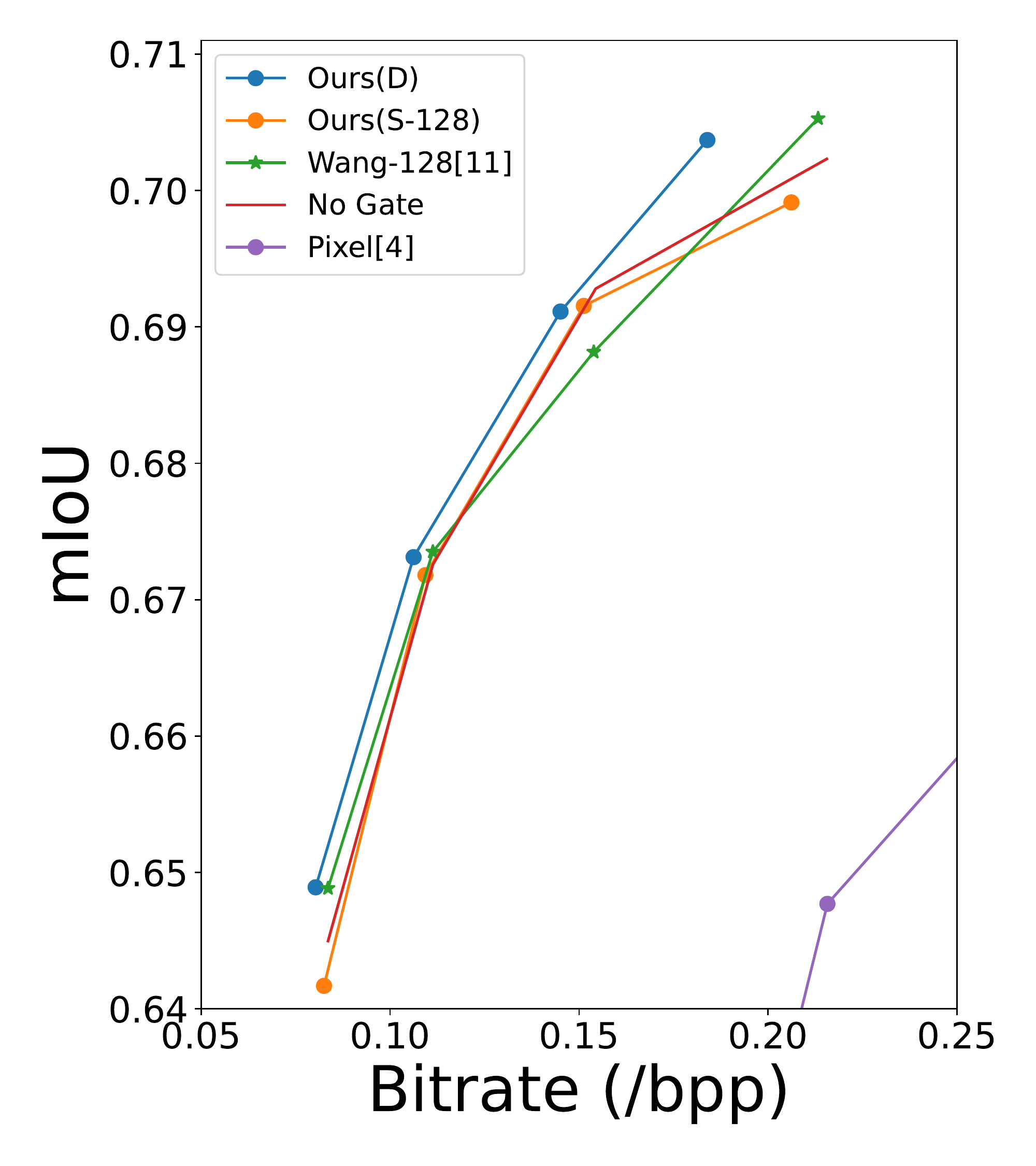}
  \end{subfigure}
  \begin{subfigure}[ht]{.24\linewidth}
    \centering\includegraphics[width=1.\linewidth]{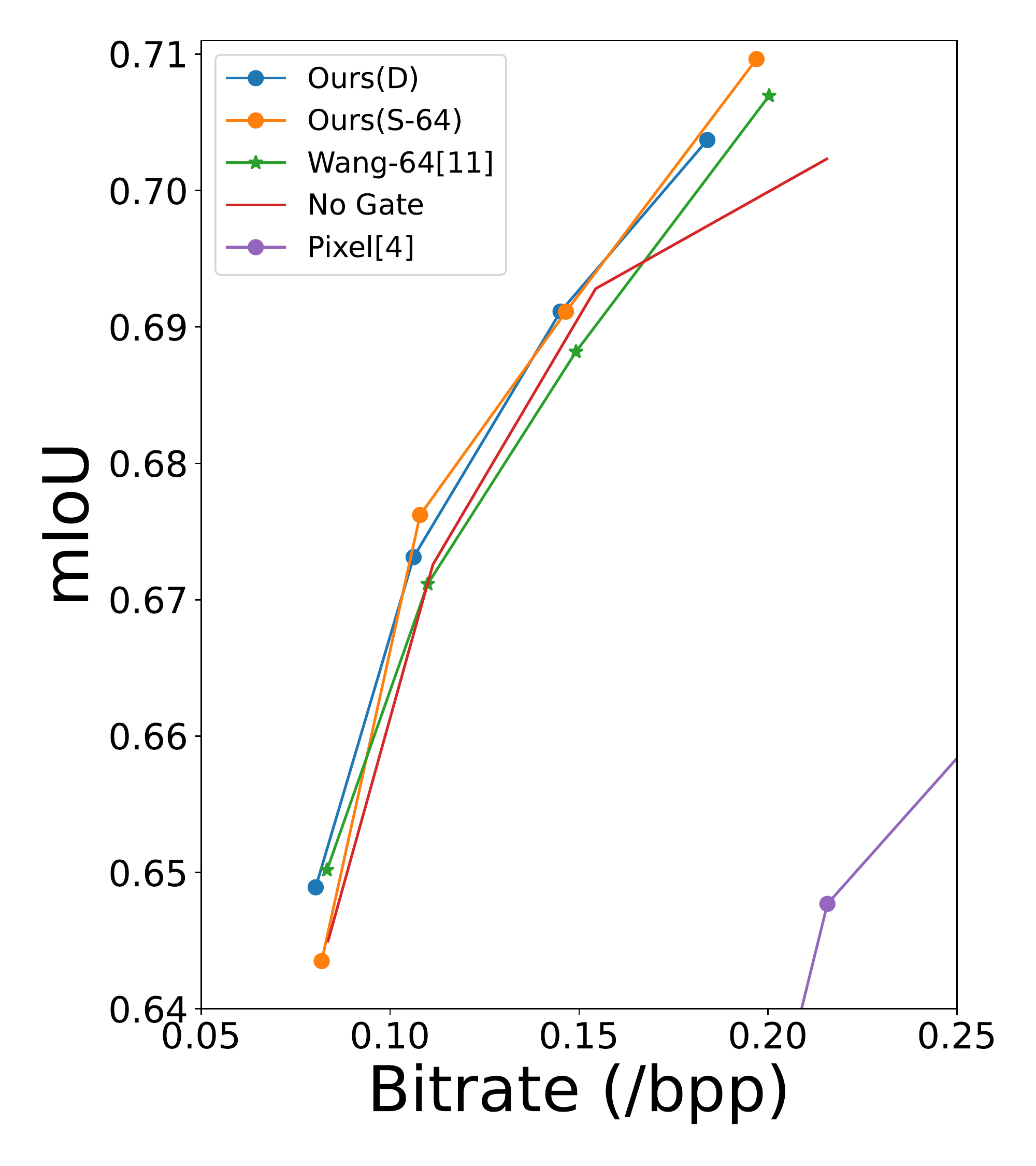}
  \end{subfigure}
  \begin{subfigure}[ht]{.24\linewidth}
    \centering\includegraphics[width=1.\linewidth]{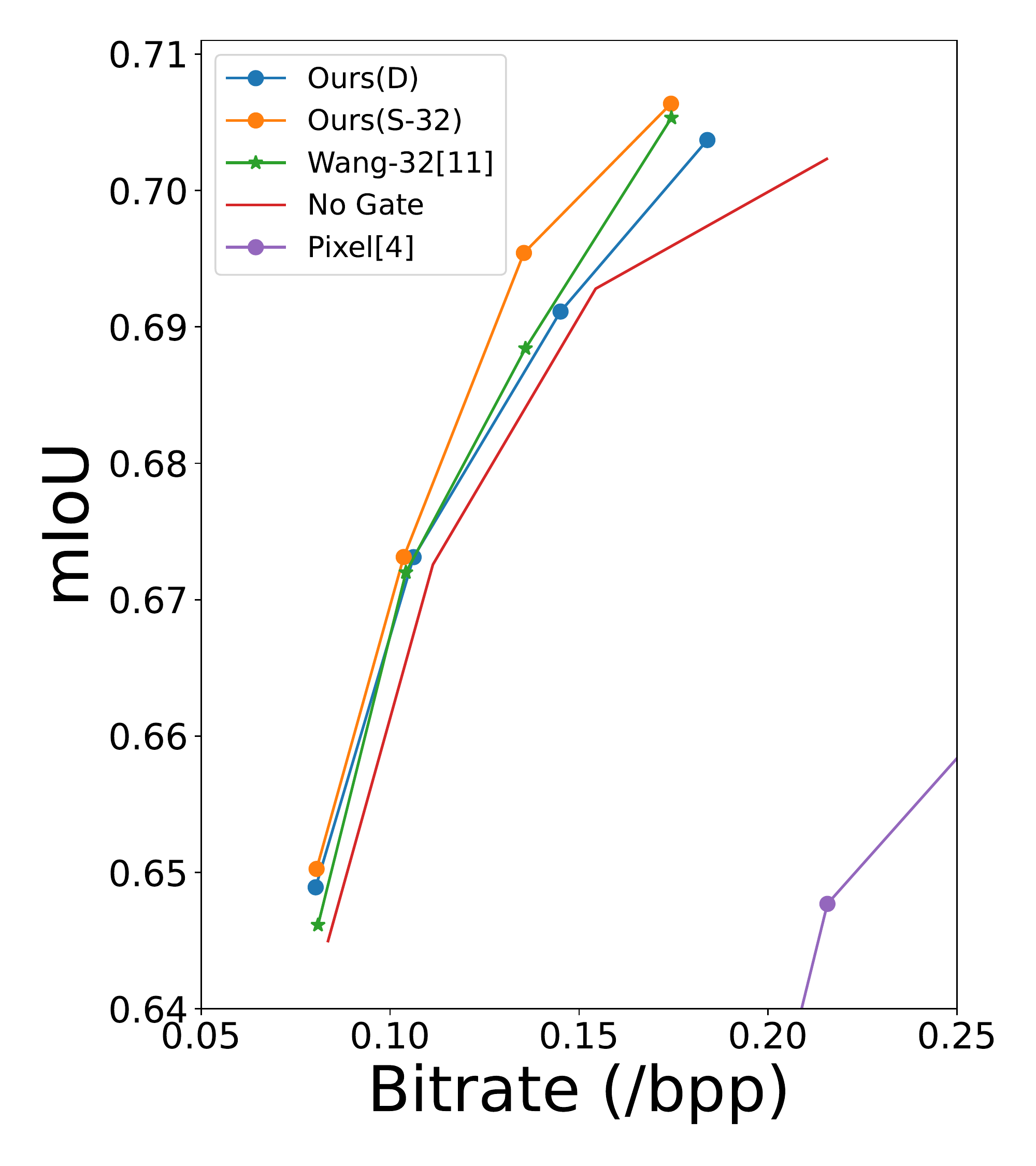}
  \end{subfigure}
  \begin{subfigure}[ht]{.24\linewidth}
    \centering\includegraphics[width=1.\linewidth]{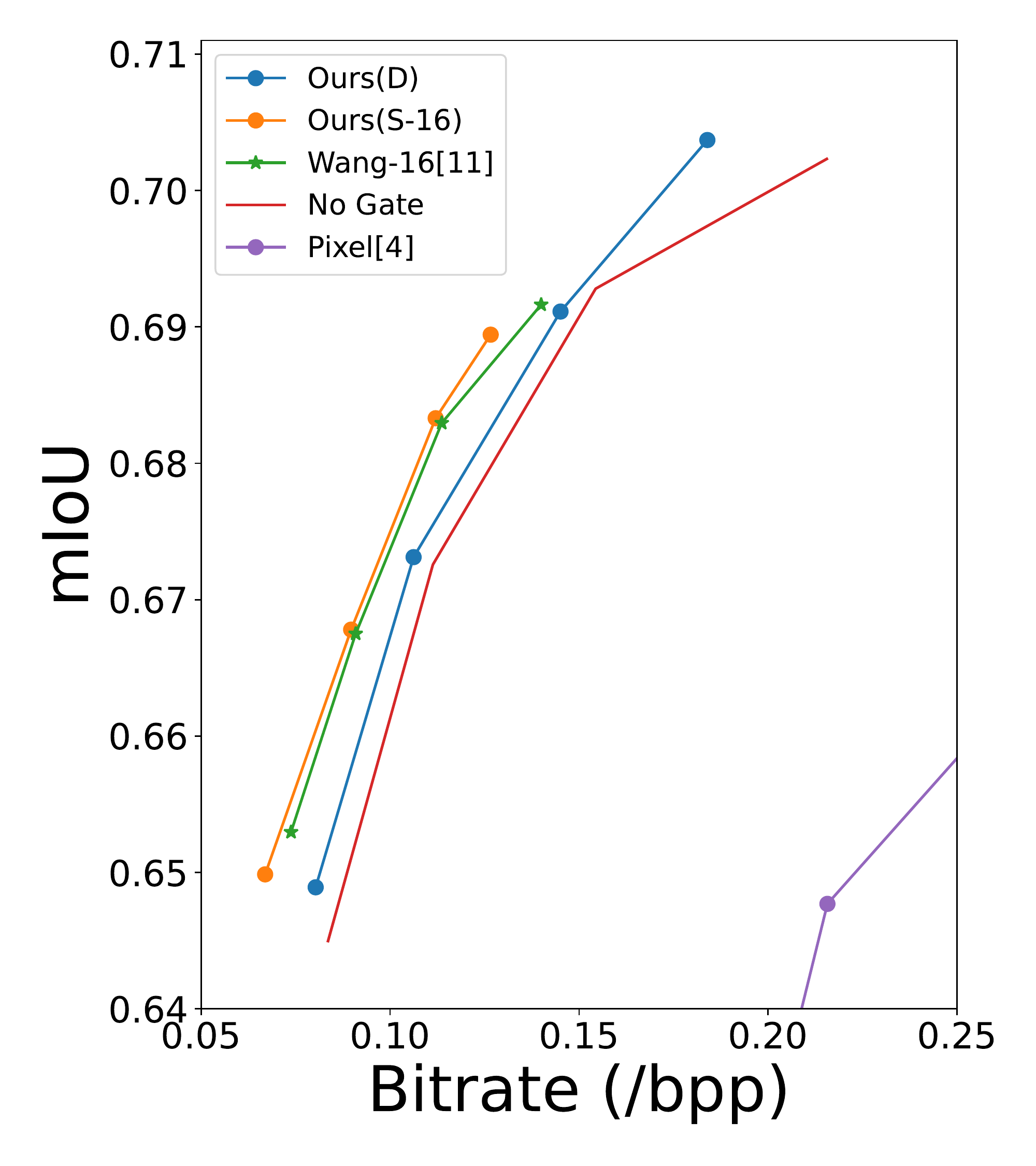}
  \end{subfigure}
  \caption{The results of training different numbers of channels of compressed domain based on Module-Deconv (Ours(D) means the dynamic method, while Ours(S-128/64/32/16) means the static method using 128/64/32/16 channels respectively, Wang-128/64/32/16 means we select 128/64/32/16 channels by using the method in \cite{wang2022learning}, "No Gate" means the compressed domain-based method without using any channel selection method, "Pixel" means the pixel domain-based method).}
  \label{fig:3}

\end{figure*}

\subsubsection{Gate Module}
The gate is designed as Fig. \ref{fig:10} shows. When we get the compressed representation $\hat{y}$, we input it to this gate module to get the selected representation $\hat{y}_{selected}$. Because in the process of channel selection, we need to sample channels, which may make the network non-differentiable. Therefore, similar to \cite{xu2020learning}, the gumbel-softmax layer is introduced for making the entire gate module differentiable.
\label{gate}

\subsubsection{Dynamic Selection}
Inputting the representation $\hat{y}$ directly into the gate module allows us to select channels dynamically. For example, we assume the size of $\hat{y}$ is $H\times W\times C$. After going through the pooling layer and the two convolutional layers, we have a tensor of size $1\times 1\times 2C$ where the first $C$ number represents the probabilities that $C$ channels of compressed representation $\hat{y}$ is sampled as 0 respectively, while the latter C numbers mean the probabilities that the $C$ channels are sampled as 1. With the final Gumbel-Softmax layer, we can get a tensor of size of $1\times 1\times C$, where each element has a value of 0 or 1, indicating that we do not select or select the corresponding channel. Finally, we multiply this tensor by $\hat{y}$ to get $\hat{y}_{selected}$. The overall flow is shown in Fig. \ref{fig:10}.
\subsubsection{Static Selection}
However, dynamic selection will make each image select different channels, which may lead to the instability of training.
Therefore, on the basis of dynamic selection, we further propose a static selection method. After the training of the whole network according to the method of dynamic selection, we use all images in the training dataset to make inferences and count the number of times each channel is dynamically selected. And then, we select the $N$ channels selected most frequently in this statistic. At last, we select these channels manually instead of using the gate module to get $\hat{y}_{selected}$.

\subsection{Transform Module}
\label{trans}
After getting $\hat{y}_{selected}$, we input it to a transform module, because the size of the compressed representation is different from that of the feature in the segmentation network, and the compressed domain has a gap with the domain in the segmentation network. To effectively bridge these gaps, and convert the compressed representation $\hat{y}_{selected}$ into the feature $\mathcal{F}$ for segmentation. We tested two different transform modules. 
\par
Refer to \cite{balle2018variational}, we firstly design the Module-Deconv which contains two transpose 
convolutional layers as Fig. \ref{fig:2}(a) shows, that is also used in \cite{liu2021}. To improve performance, according to \cite{cheng2020learned}, Module-Res is designed as Fig. \ref{fig:2}(b) shows. The residual block and PixelShuffle are used for improving performance.

\section{Experiments}
\subsection{Experiment Setting}
For the image compression network, we use the hyperprior model in \cite{balle2018variational}. For semantic segmentation network, we use DeepLab v3 \cite{chen2017rethinking}. The Cityscapes dataset \cite{Cordts2016Cityscapes} is used as the training set and the test set. Adam is chosen as the optimizer. We totally train this network by 400 epochs while the initial learning rate is set as 0.001 and drops to 1e-4 after 300 epochs. All experiments are based on a GeForce RTX 3090 GPU with 24 GB RAM.
\begin{table}[]
\caption{The comparison between the compressed domain-based method without channel selection and the pixel domain-based method in the same bitrate cases.}
 \setlength{\tabcolsep}{0.5mm}{\begin{tabular}{l|l|l|l|l}
\hline
\textbf{Bitrate(/bpp)}                                                          & \textbf{0.084}                                           & \textbf{0.111}                                           & \textbf{0.154}                                           & \textbf{0.216}                                       \\ \hline\hline
\begin{tabular}[c]{@{}l@{}}\textbf{mIoU(Pixel} \\ \textbf{domain-based)}\end{tabular}       & 0.423                                                    & 0.514                                                    & 0.579                                                    & 0.648                                                   \\ \hline
\begin{tabular}[c]{@{}l@{}}\textbf{mIoU(Compressed} \\ \textbf{domain-based)}\end{tabular} & \begin{tabular}[c]{@{}l@{}}0.645\\ \color{blue}(52.5\%$\uparrow$)\end{tabular} & \begin{tabular}[c]{@{}l@{}}0.673\\ \color{blue}(30.9\%$\uparrow$)\end{tabular} & \begin{tabular}[c]{@{}l@{}}0.693\\ \color{blue}(19.7\%$\uparrow$)\end{tabular} & \begin{tabular}[c]{@{}l@{}}0.702\\ \color{blue}(8.3\%$\uparrow$)\end{tabular} \\ \hline
\end{tabular}}
\label{tab:1}
\end{table}
\subsection{Rate-mIou Comparison}
We test dynamic and static selection methods based on Module-Deconv in this part. In the experiments of the static selection method, we test the cases when the number of selected channels is 128, 64, 32, and 16, respectively. 
\par
To prove that redundancy does exist in the compressed representation, we compare our method with the case without using any channel selection method. As the Fig. \ref{fig:3} shows, compared with that case, we can save up to 31.8\% bitrates.
\begin{figure}[t]
	\centering
	\vspace{-0.6cm}  
    \setlength{\abovecaptionskip}{-0.1cm}   

	\includegraphics[scale=0.4]{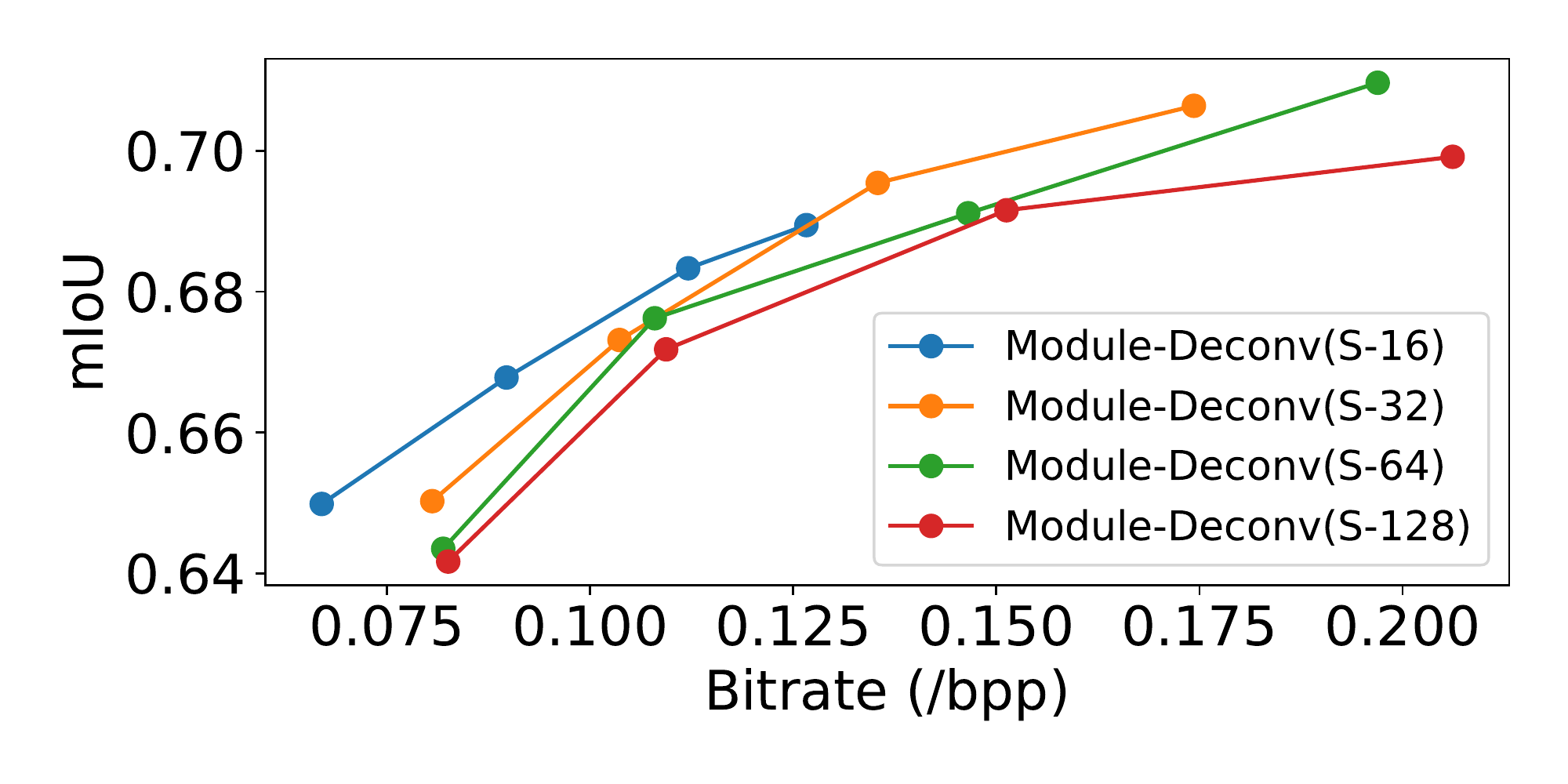}
	\caption{The comparison of various channels.
	}
	\label{fig:4}
\end{figure}
\begin{table}[]
\centering
\caption{The performance and MACs (Multiply-Add Cumulation) of various transform modules.}
\setlength{\tabcolsep}{2mm}\begin{tabular}{l|l|l}
\hline
\textbf{Module Name}       & \begin{tabular}[c]{@{}l@{}}\textbf{Module}\textbf{-Deconv} \cite{liu2021}\end{tabular}  & \begin{tabular}[c]{@{}l@{}}\textbf{Module}\textbf{-Res} \end{tabular} \\ \hline\hline
\textbf{MACs(/G)}   & \textbf{94}                & 202                      \\ \hline\hline
\textbf{mIoU(0.067 bpp)} & 0.6502              & \textbf{0.6534}             \\ 
\textbf{mIoU(0.104 bpp)} & 0.6785             & \textbf{0.6836}                  \\ \hline
\end{tabular}
\label{tab:2}
\end{table}
\par
To verify the effectiveness of our method, we firstly compare our method with the pixel domain-based method whose compression network is also \cite{balle2018variational}. As Table \ref{tab:1} shows, our compressed domain-based method is superior to the pixel domain-based method.
In addition, we compare our method with a state-of-the-art compressed domain-based method \cite{wang2022learning}. Here, we reproduce \cite{wang2022learning} by using the same image compression network in our method.
As Fig. \ref{fig:3} shows, regardless of the number of channels selected, our static method always outperforms \cite{wang2022learning}. For example, if the number of selected channels is 32, we can 
save up to 15.8\% bitrates compared with \cite{wang2022learning} and save up to about 83.6\% compared with the pixel domain-based method. This shows that our method can select more appropriate channels for segmentation tasks.
The static method has a better rate-mIoU when the selected channels are less than 128, which suggests that static method is superior to the dynamic method.
\subsection{The Rate-PSNR Comparsion}
Since the weights of image compression network are frozen during training process, the rate-PSNR is the same as the results of \cite{balle2018variational}.
Also, we reproduce the channel selection method \cite{wang2022learning} by using the same image compression network as ours, which means that we have the same rate-PSNR performance. 

\subsection{The Effect of Various Channels}
In Fig. \ref{fig:4}, we test the cases of various channels. As we can see, compared with selecting 64 or 128 channels, selecting 16 or 32 channels can get better rate-mIou which means we reduced the redundancy in the compressed representation.
\subsection{The Effect of Various Transform Modules}
In this paper, we explore two different transform modules. As Table \ref{tab:2} shows, 
Module-Res can get a better rate-mIoU in both bitrates while Module-Deconv perform slightly worse but with half the complexity compared with Module-Res.
\subsection{Inference Time}
To check the efficiency of our method, we test the inference time based on CityScapes dataset, as Table. \ref{tab:3} shows. "No gate" means the compressed domain-based method without using any channel selection method. As we can see, compared with the pixel domain-based method, compressed domain-based methods can save a lot of time. Moreover, with dynamic selection or static selection we can save much more time, especially when selected channels are 16, we save about 44.8\% of the time.

\begin{table}[]
\caption{The inference time for semantic segmentation based on CityScapes dataset.}
\setlength{\tabcolsep}{1.6mm}\begin{tabular}{ll|l|l}
\hline
\multicolumn{2}{l|}{\textbf{Model}}                                                &  \begin{tabular}[c]{@{}l@{}}\textbf{Encoding} \\\textbf{time(/s)}\end{tabular} & \begin{tabular}[c]{@{}l@{}}\textbf{Decoding + }\\ \textbf{task time(/s)}\end{tabular} \\ \hline\hline
\multicolumn{2}{l|}{\textbf{Pixel Domain}}                                         & 0.142                  & 0.320                         \\ \hline\hline
\multicolumn{2}{l|}{\textbf{Ours(Dynamic Selection)}}                                    & 0.090                  & 0.217                         \\ \hline\hline
\multicolumn{1}{l|}{\multirow{5}{*}{{\begin{tabular}[c]{@{}l@{}}\textbf{Ours}\\ \textbf{(Static Selection)}\end{tabular}}}} & \textbf{16 channels}      & 0.060                  & 0.195                         \\ \cline{2-4} 
\multicolumn{1}{l|}{}                                           & \textbf{32 channels}      & 0.067                  & 0.198                        \\ \cline{2-4} 
\multicolumn{1}{l|}{}                                           & \textbf{64 channels}      & 0.082                  & 0.208                         \\ \cline{2-4} 
\multicolumn{1}{l|}{}                                           & \textbf{128 channels}     & 0.111                  & 0.226                         \\ \cline{2-4} 
\multicolumn{1}{l|}{}                                           & \textbf{No gate} & 0.139                  & 0.247                        \\ \hline
\end{tabular}

\label{tab:3}
\end{table}
\subsection{Scalability for Multi-tasks}
Since the weights of image compression network are frozen, the scheme is scalable for multi-tasks. We explore the potential scalability by testing object detection with our dynamic selection method based on  PASCAL VOC 07 dataset. Faster-rcnn \cite{ren2015faster} is used as task network. The results show that we save up to about 30\% bitrates to achieve similar average precision compared with the pixel domain-based method. It's our future work to further explore the scalability.
\section{Conclusions}
In this paper, we propose a method based on the compressed domain to improve segmentation tasks. We firstly proposed a gate module to learn adaptively which channels in the compressed representations are suitable for segmentation tasks. After that, the static selection of a specified number of channels and dynamic selection based on the gate module are studied. These two kinds of selection help us successfully reduce the redundancy of the compressed representations. And then, two different transform modules are introduced for transforming the representation in the compressed domain to the feature in the segmentation network. At last, we do experiments based on the Cityscapes dataset, and the results of our experiments prove that our model can achieve better rate-mIoU and faster inference compared with the previous work.

\small
\bibliographystyle{IEEEbib}
\bibliography{main}

\end{document}